\documentclass[conference]{IEEEtran}
\IEEEoverridecommandlockouts
\usepackage{cite}
\usepackage{amsmath,amssymb,amsfonts}
\usepackage{graphicx}
\usepackage{textcomp}
\usepackage{xcolor}

\usepackage[ruled, linesnumbered, vlined]{algorithm2e} 
\usepackage{longtable}
\usepackage{multirow}
\usepackage{diagbox}
\usepackage{pgfplots}
\usepackage{caption}
\usepackage{tikz}

\SetKwInOut{Input}{Input}  
\SetKwInput{Output}{Output}  

\makeatletter
\newcommand{\nosemic}{\renewcommand{\@endalgocfline}{\relax}}
\newcommand{\dosemic}{\renewcommand{\@endalgocfline}{\algocf@endline}}

\newcommand{\pushline}{}
\let\oldnl\nl
\newcommand{\nonl}{\renewcommand{\nl}{\let\nl\oldnl}}

\begin{document}

\title{Federated Learning for Non-IID Data via Client Variance Reduction and Adaptive Server Update}

\author{\IEEEauthorblockN{Hiep Nguyen}
\IEEEauthorblockA{
\textit{HCL Technologies, Vietnam}\\
hiep\_nguyen@hcl.com}
\and
\IEEEauthorblockN{Lam Phan}
\IEEEauthorblockA{
\textit{HCL Technologies, Vietnam}\\
lam.phan@hcl.com}
\and
\IEEEauthorblockN{Harikrishna Warrier}
\IEEEauthorblockA{
\textit{HCL Technologies, India}\\
harikrishna.w@hcl.com}
\and
\IEEEauthorblockN{Yogesh Gupta}
\IEEEauthorblockA{
\textit{HCL Technologies, India}\\
yogeshg@hcl.com}
}

\maketitle

\begin{abstract}
 Federated learning (FL) is an emerging technique used to collaboratively train a global machine learning model while keeping the data localized on the user devices. The main obstacle to FL’s practical implementation is the Non-Independent and Identical (Non-IID) data distribution across users, which slows convergence and degrades performance. To tackle this fundamental issue, we propose a method (ComFed) that enhances the whole training process on both the client and server sides. The key idea of ComFed is to simultaneously utilize client-variance reduction techniques to facilitate server aggregation and global adaptive update techniques to accelerate learning. Our experiments on the CIFAR-10 classification task show that ComFed can improve state-of-the-art algorithms dedicated to Non-IID data. 
\end{abstract}


\section{Introduction}
\begin{figure}
    \centering
    \includegraphics[scale=0.45]{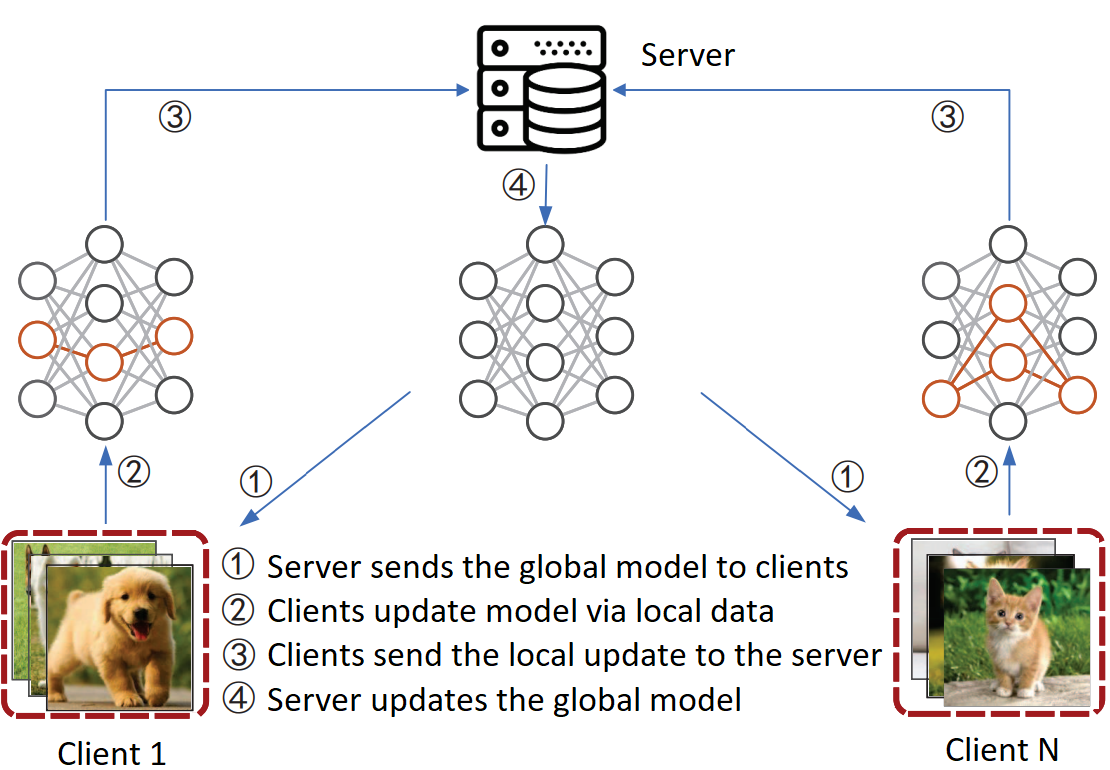}
    \caption{The pipeline of Federated learning}
    \label{fig:FL}
\end{figure}

Federated learning is a privacy-preserving distributed machine learning mechanism that allows numerous clients (e.g., edge devices or organizations) to collaboratively train a shared global model without exchanging their private data.
In each round of FL, every participating client receives an initial model from a central server, trains the model using its local dataset, and then sends back the local update to the server 
to form a new global model for the next round (see Figure~\ref{fig:FL}).

A key challenge for Federated learning is the heterogeneity of data distribution across the network. In reality, the data can vary dramatically across clients concerning data quantity, label distribution, and feature distribution. Under heterogeneous federated data distribution, local models move away from globally optimal models and become divergent. Aggregating divergent local models slows down the convergence rate and degrades the accuracy of the global model~\cite{FedAvgM,NIID_silos,FedAvg_NIID,FedAvg_NIID_shareddata}.

The most popular algorithm for Federated learning is FedAvg~\cite{FedAvg}.  Despite its demonstrated success in some applications, FedAvg does not fully tackle the underlying problem of data heterogeneity. 
Various FL algorithms have been proposed in the literature to mitigate the negative impact of the data heterogeneity and facilitate the convergence of Federated learning. They mainly employ one of two techniques: client-variance reduction or adaptive global model update. The former strategies aim at pulling local models towards the global model, making the server model aggregation easier. In contrast, the latter techniques aim at dampening the global model update oscillations, accelerating learning.

\textbf{Client-variance reduction}. The variance here refers to the deviation in weights, update directions or feature representations among local models. 
FedProx~\cite{FedProx} adopts a regularization term into the local objective function to limit the distance between the local models and the global model. 
Moon~\cite{Moon} uses a contrastive loss to enforce the agreement between the feature representation learned by the server and client models. 
VLR-SGD~\cite{VLRSGD} adjusts local updates based on the difference in weights between the server and client models.
Scaffold~\cite{Scaffold} corrects local updates via control variates that estimate the update direction of the server model and client models. 
FedDC~\cite{FedDC}, a combination of FedProx and Scaffold, applies regularization and local update correction at the same time. 
Mime~\cite{Mime} embeds a server momentum into every client update to mimic the behavior of the centralized algorithm running on Non-IID data.  
FedADC~\cite{FedAdc} normalizes server momentum before integrating it into local updates. 
To mitigate the negative impact caused by heterogeneous local updates, FedNova~\cite{FedNova} normalizes them before aggregating.

\textbf{Adaptive global model update}.
SlowMo~\cite{SlowMo} and FedAvgM~\cite{FedAvgM} use a server momentum to dampen oscillation. 
They modify the server model based on the cumulative update history rather than just the current average local update that can vary significantly across rounds. 
The methods proposed in~\cite{Adaptive}, including FedAdam, FedAdagrad, and FedYogi, apply adaptive server optimizers that support both server-side momentum techniques and adaptive global learning rates. They have been experimentally demonstrated to accelerate the convergence rate of Federated learning in Non-IID data settings.

Previous studies in the literature frequently apply one of these two techniques alone to address the data heterogeneity. Algorithms based on client-variance reduction focus on learning techniques on the client side while using a naive global update strategy. On the contrary, algorithms based on global adaptive updates concentrate on improving the learning on the server side while using a simple Sgd-based client learning strategy. In other words, FL algorithms for Non-IID data are primarily concerned with optimizing the training process on one side, either client-side or server-side, resulting in limited performance. This restriction motivates us to propose ComFed, a method that simultaneously applies the two techniques to enhance the whole training process on both sides. ComFed would leverage the robustness of both techniques against Non-IID data distribution.

We investigate the effectiveness of 16 combinations between the client learning strategies of FedAvg, FedProx, Scaffold, and FedNova with the server update strategies of FedAvg, FedAdam, FedAdagrad, and FedYogi. To the best of our knowledge, this work provides the first study on systematically understanding the impact of various combinations between client-variance reduction techniques and server update techniques under Non-IID settings. The experimental results show that our ComFed could improve the test accuracy in some cases. 
The ideal case for applying our approach is when both individual techniques outperform FedAvg.

\section{Preliminaries}
Consider a Federated learning system across N users, each with its local data set $D_i$ of $n_i = |D_i|$ samples, for $i=1,\ldots, N$. $p_i = |D_i|/ \sum_{i=j}^{N} |D_i|$ denotes the data ratio of user $i$. We aim to learn a machine learning model $w$ over the distributed data set $D$ that satisfies the following objective function:
\begin{center}
    $\underset{w}{\mathrm{argmin}} [ F(w) = \sum_{i=1}^{N} p_i F_i(w) ] $
\end{center}

where $F_i(w) = \frac{1}{n_i} \sum_{x \in D_i}{f_i (w,x)}$ is the local objective function of client $i$, and $f_i$ is the arbitrary loss function defined by the learning model $w$ and data sample $x$ in $D_i$. 
This federated optimization problem is solved by achieving local objective functions iteratively, as described in Algorithm~\ref{alg:FL}. At the beginning of each round $t$, the server broadcasts its current global model to a subset of randomly selected clients with a sampling ratio $C$ (line~\ref{alg:FL:broadcast}). Each round consists of two phases: local training and global update.

In the local training phase, each client $i$ iteratively updates its local model $w_i$  (line~\ref{alg:FL:lupdate}) and then sends back the local update $\Delta_i$, defined as the total modification of the local model in the round, to the server (line~\ref{alg:FL:send_back}).  $w_i$ is initialized to the global model received at the beginning of the round (line~\ref{alg:FL:init}). Under the Stochastic Gradient Descent (Sgd) rule, the local model is simply changed by $-\eta g_i$ for each data batch $b_j$, where $\eta$ denotes the local learning rate, and $g_i$ denotes the mini-batch gradient. In the general case, the local model is modified by 
an amount proportional to $g_i$ and $\eta$ (line~\ref{alg:FL:lupdate}), which depends on the used FL algorithm. 

In the server update phase, the server aggregates local updates $\Delta_i$ (line~\ref{alg:FL:aggregation})
and modifies the global model based on the aggregated update to produce a new global model for the next round (line~\ref{alg:FL:gupdate}). The most straightforward strategy is to use the averaging aggregation and modify the global model by the average local update. In the general case, the global model is modified by an amount proportional to the aggregation of local updates and the global learning rate $\eta_g$. 

\begin{algorithm}
\caption{Generalized FL algorithm\label{alg:FL}}


\Input{number of clients N, client selection ratio C,\\
number of rounds T, number of local epochs E,\\
local learning rate $\eta$, global learning rate $\eta _g$
}
\Output{global model $w$}
\BlankLine

\For{round $t = 1, \ldots, T$}{
    Select a subset $S$ of clients, $|S|=C*N$ \\
    Broadcast global model $w$ \label{alg:FL:broadcast} to clients in $S$\\
    
    \pushline\dosemic\nonl \textbf{\textit{Local training phase}} \\
    
    \For{client $i$ in $S$ \textbf{\textup{in parallel}}}{
    $w_{i} \leftarrow w$   \label{alg:FL:init}\\
    \For{epoch $e=1,\ldots, E$ \label{alg:FL:loopE}}{
    \For{batch $b_j$ of $D_i$ \label{alg:FL:loop_batch}}{
        compute mini-batch gradient $g_i \leftarrow \sum_{x\in b_j}{\nabla f_i(w_i, x)}$   \label{alg:FL:grad} \\
        update $w_i \leftarrow$ $upd_c$ ($g_i, \eta$)      \label{alg:FL:lupdate}
    }
    }
    Send local update $\Delta_i \leftarrow w_i - w$ to the server         \label{alg:FL:send_back}
    }

    \pushline\dosemic\nonl \textbf{\textit{Server update phase}} \\

    Aggregate local updates $\Delta \leftarrow agg(\{\Delta_{i \in S}\})$ \label{alg:FL:aggregation}\\
    Update global model $w \leftarrow$ $upd_s$($\Delta, \eta_g$) \label{alg:FL:gupdate}
        
}
\end{algorithm}

\begin{table*}

\begin{center}
\begin{tabular}{|l||l|l|l|l|}
\hline

        &\multicolumn{2}{c|}{Local training phase}  &	\multicolumn{2}{c|}{Global update phase} \\
\cline{2-5}
Algorithm        & Local loss $f_i$    &	Local update $upd_c$   &	Aggregation   $agg$  &	Global update $upd_s$ \\
\hline
FedAvg~\cite{FedAvg}	&	log loss	&	\shortstack[l]{Sgd: \\ $w_i = w_i -\eta g_i$}	&	 \shortstack[l]{averaging: \\ $\Delta = \frac{1}{|S|} \sum_{i \in S}{\Delta_i}$}	&	\shortstack[l]{pseudo-Sgd with $\eta_g = 1$: \\$w = w + \Delta_i$  }	\\ \hline
Moon~\cite{Moon}	&	contrastive loss	&	\multicolumn{1}{|c|}-	&	\multicolumn{1}{|c|}-	&	\multicolumn{1}{|c|}-	\\ \hline
FedProx~\cite{FedProx}	&	regularized	loss &	\multicolumn{1}{|c|}-	&	\multicolumn{1}{|c|}-	&	\multicolumn{1}{|c|}-	\\ \hline
FedAdc~\cite{FedAdc}	&	\multicolumn{1}{|c|}-	&	Sgd$_m$ 	&	\multicolumn{1}{|c|}-	& server momentum	\\ \hline
Mime~\cite{Mime}	&	\multicolumn{1}{|c|}-	&	\shortstack[l]{corrected Sgd}	&	\multicolumn{1}{|c|}-	&	\multicolumn{1}{|c|}-	\\ \hline
VRL$-$SGD~\cite{VLRSGD}	&	\multicolumn{1}{|c|}-	&	corrected Sgd	&	\multicolumn{1}{|c|}-	&	\multicolumn{1}{|c|}-	\\ \hline
Scaffold~\cite{Scaffold}	&	\multicolumn{1}{|c|}-	&	corrected Sgd	&	\multicolumn{1}{|c|}-	&	\multicolumn{1}{|c|}-	\\ \hline
FedDC~\cite{FedDC}	&	regularized	loss &	corrected Sgd	&	\multicolumn{1}{|c|}-	&	\multicolumn{1}{|c|}-	\\ \hline

FedNova~\cite{FedNova}	&	\multicolumn{1}{|c|}-	&	\multicolumn{1}{|c|}-	&	normalized averaging	&	\multicolumn{1}{|c|}-	\\ \hline	
qFFL~\cite{qFFL}	&	\multicolumn{1}{|c|}-	&	\multicolumn{1}{|c|}-	&	exponentially averaging 	&	\multicolumn{1}{|c|}-	\\ \hline
SlowMo~\cite{SlowMo}	&	\multicolumn{1}{|c|}-	&	\multicolumn{1}{|c|}-	&	\multicolumn{1}{|c|}-	&	server momentum, $\eta_g = 1$	\\ \hline
FedAvgM~\cite{FedAvgM}	&	\multicolumn{1}{|c|}-	&	\multicolumn{1}{|c|}-	&	\multicolumn{1}{|c|}-	&	server momentum, $\eta_g = 1$	\\ \hline
FedAdam~\cite{Adaptive}	&	\multicolumn{1}{|c|}-	&	\multicolumn{1}{|c|}-	&	\multicolumn{1}{|c|}-	&	 \\	\cline{1-4}
FedAdagrad~\cite{Adaptive}	&	\multicolumn{1}{|c|}-	&	\multicolumn{1}{|c|}-	&	\multicolumn{1}{|c|}-	&	server momentum, adaptive $\eta_g$	\\ \cline{1-4}
FedYogi~\cite{Adaptive}	&	\multicolumn{1}{|c|}-	&	\multicolumn{1}{|c|}-	&	\multicolumn{1}{|c|}-	&		\\ 

\hline
\end{tabular}

\end{center}
\caption{The characteristics of FL algorithms. Symbol '-' implies the same technique as FedAvg. Sgd$_m$ means Sgd with momentum. 
\label{Table_summary}}
\end{table*}

Each FL algorithm is characterized by 4 elements: local loss function $f_i$, local update formula $upd_c$, aggregation method $agg$, and global update formula $upd_s$. 
Many adjustments related to one of these elements have been proposed in the literature to address the issue of Non-ID data (see {Table~\ref{Table_summary}). 
The first two elements are associated with the local training phase and the last two with the global update phase. 
Techniques that focus on improving the local training phase attempt to reduce the variance among clients. On the contrary, techniques that improve the global update phase aim at preventing oscillation and accelerating learning. 
The next section will briefly present FedAvg, the de facto standard algorithm, and the algorithms dedicated to Non-IID data that will be evaluated throughout the paper and integrated into our ComFed.

\section{Baseline algorithms \label{sec_baseline}}


\subsection{FedAvg}

FedAvg~\cite{FedAvg} uses a weighted averaging aggregator, with weights proportional to client data ratios. FedAvg updates client models according to the Sgd rule and updates the server model based on the average local update, as follows:

\begin{itemize}
\item Local update: $w_i = w_i -\eta g_i$  (refer to line~\ref{alg:FL:lupdate} Algorithm~\ref{alg:FL})
\item Aggregation: $\Delta = \sum_{i \in S}{p'_i \Delta_i}$ (refer to line~\ref{alg:FL:aggregation})
\item Global update: $w = w + \Delta$ (refer to line~\ref{alg:FL:gupdate})
\end{itemize}

where $p'_i= \frac{n_i}{\sum_{i \in S} n_i}$ is the data percentage of client $i$ over the clients selected in the round.  
We can consider that FedAvg updates the server model by a pseudo-Sgd on the gradient $- \Delta$ with a learning rate $\eta_g = 1$.
Algorithm~\ref{alg:FL} performs FedAvg when  
it uses the averaging aggregation for the $agg$ function in line~\ref{alg:FL:aggregation} and updates the server model by the average of local updates with $\eta_g=1$ (line~\ref{alg:FL:gupdate}).

\subsection{FedProx}
FedProx improves FedAvg by modifying the local objective function. It adds an l2-regularization on the distance between the global model learned in the previous round and the current local model. 
 Instead of just minimizing the local loss function $f_i$, client $i$ minimizes the dissimilarity between its local model $w_i$ and the global model as follows:
\begin{center}
  minimize $f_i + \frac{\mu}{2}
  \lVert {w - w_i} \rVert^2$
\end{center}

 where $\mu \geq 0$ is a parameter that controls the impact of the proximal term. FedAvg is a particular case of FedProx with $\mu = 0$. The role of this proximal term is to pull local models towards the global model, and as a result, the divergence among clients reduces. FedProx demonstrates more stable and accurate convergence over FedAvg.  
 Algorithm~\ref{alg:FL} performs FedProx when it replaces the local objective function $f_i$ in line~\ref{alg:FL:grad} with $f_i + \frac{\mu}{2} \lVert {w - w_i} \rVert^2$.

\subsection{Scaffold\label{subsec_Scaffold}}
Scaffold\cite{Scaffold} uses control variates $c, ci$ to estimate the model update direction of the server and client $i$. It proposes two methods for updating $c_i$ and uses the average rule for aggregating $c$:

\begin{itemize}
    \item $c_i = \nabla f_i(w,D_i)$: $c_i$ is the gradient of the global model at the local data. 
    \item $c_i = c_i - c + \frac{1}{K\eta} (w - w_i)$, where $K$ is the number of local steps corresponding to the number of local data batches visited per round. 
    \item $c \leftarrow c + \frac{1}{|S|} \sum_{i \in S} p_i' \Delta c_i$, where $\Delta c_i$ denotes the modification of $c_i$ in the current round. 
\end{itemize}

Control variates are exchanged between the clients and server along with model weights (Algorithm \ref{alg:FL}, lines~\ref{alg:FL:broadcast} and~\ref{alg:FL:send_back}) and maintained across rounds. 
The difference between the two update directions, $c - c_i$, evaluates the client drift and is used to correct the local update  as follows: 
\begin{center}
$w_i = w_i - \eta (g_i -c_i + c)$ (Algorithm \ref{alg:FL} line~\ref{alg:FL:lupdate})
\end{center}
The correction term $c-ci$ ensures that the local and global models are updated in close directions to reduce client drift. Compared to FedAvg, Scaffold doubles the computational and communication cost due to the maintenance of control variates across rounds. 



\subsection{FedNova}

FedNova modifies FedAvg by using a normalized aggregation instead of the simple averaging aggregation. 
 It considers that clients can perform different numbers of local updates, employ different local optimizers and learn at different learning rates, resulting in heterogeneous local progress.
Clients with larger local updates will significantly impact the global update. To avoid a bias in the global update, FedNova normalizes then re-scales the accumulated local update $\Delta_i$ before averaging:
\[
\Delta = \gamma \sum_{i \in S}{ p'_i  \frac{\Delta_i}{\lVert a_i \rVert_1} }
\]
where $\gamma$ is a scaling factor 
, $a_i = [a_{i1}, a_{i2}, \ldots]$ is a vector of coefficients that defines how client $i$ accumulates its mini-gradients $g_{ij}$ 
in the round : $\Delta_i = \sum_{b_j \in D_i}{a_{ij}g_{ij}}$.  
The calculation formula for $\lVert a_i \rVert_1$ varies depending on local optimizers. In the case of FedAvg using local Sgd, $a_i = [1,1, \ldots, 1]$ and $\lVert a_i \rVert_1 = \frac{1}{\tau_i} $, where $\tau_i$ is the number of local steps in the current round. In this case, the normalized gradient $\frac{\Delta_i}{\lVert a_i \rVert_1} = \frac{\Delta_i}{\tau_i}$ is simply an average of all gradients within the current round.  
FedNova can be done by Algorithm~\ref{alg:FL} when using a normalized aggregation in line~\ref{alg:FL:aggregation} and doing some extra work: maintaining $a_i$ after each local update at line~\ref{alg:FL:lupdate} and sending it back to the server with model weights at line~\ref{alg:FL:send_back}.

\subsection{FedAdam, FedAdagrad, FedYogi}

Paper~\cite{Adaptive} proposes a group of 3 algorithms, including FedAdam, FedAdagrad, and FedYogi, which modify FedAvg in the global update step by respectively applying the server-side adaptive optimizers  Adam, Adagrad, and Yogi as follows:

\vspace{1em}

$m = \beta_1 m +  (1-\beta_1)\Delta$ 

$v = v + \Delta^2$ \ \ \ \ \ \ \ \ \ \ \ \ \ \ \ \ \ \ \ \ \ \ \ \ \ \ \ (FedAdagrad) 

$v = v - (1-\beta_2) \Delta^2$ sign$(v - \Delta^2)$ \ \    (FedYogi)  \ \ \ \ \ \  \ \ \ \ \ \ \ (1)

$v = \beta_2 v - (1-\beta_2) \Delta^2$ \ \ \ \ \ \ \ \ \ \ \ \ \ \ (FedAdam)

$w = \beta_1 w +  \frac{\eta_g}{\sqrt{v} + \varepsilon} m$

\vspace{1em}

where $m$ and $v$ are the server-side momentums, $\beta_1$ and $\beta_2$ are two parameters, and $\varepsilon$ is a minimal value to avoid division by zero in the global update formula. 
The server treats the aggregated local update $\Delta$ as a pseudo-gradient 
and utilizes it to compute the global momentum $m$.
On the one hand, the server model is updated based on the accumulative update history rather than just the current average local update that can highly fluctuate across rounds. On the other hand, the learning rate  adaptively changes during the training process as $\frac{\eta_g}{\sqrt{v}+\epsilon}$. 
So, these adaptive optimizers support server momentums and adaptive learning rates, allowing dampening oscillations and controlling convergence rates.
Paper~\cite{Adaptive} shows that using adaptive server optimizers offers substantial improvements over FedAvg.
All three algorithms can be executed by Algorithm~\ref{alg:FL} when using Adam, Adagrad, or Yogi optimizers, respectively, in line~\ref{alg:FL:gupdate}.

\section{Proposed approach}

Inspired by the demonstrated success of the algorithms presented in Section~\ref{sec_baseline} for Non-IID data settings, we propose a new method called ComFed.
ComFed uses client-variance reduction techniques to reduce the gap among clients, making server model aggregation easier. Simultaneously, ComFed applies server-side adaptive update techniques to dampen oscillations, help the global gradient vector point to the right direction, and take more straightforward paths to the global optimum. Unlikely state-of-the-art algorithms, ComFed enhances the whole training process on both the client and server sides. Intuitively, ComFed can benefit from the robustness of both kinds of techniques against Non-IID data distributions.

We consider three client-variance reduction techniques used in FedProx, Scaffold, and Fedorova. They are to regularize local objectives by a proximal term (briefly called Prox), correct local updates with control variates (called Scaf), and normalize local updates (called Nova). 
Prox, Scaf, and Nova techniques adjust three steps: mini-batch gradient calculation, local update, and aggregation, respectively. 
To simplify the definition of our algorithms, we group these three steps into a stage.
Each learning mechanism associated with this stage, denoted by $opt_c$, specifies how to calculate mini-batch gradients, update local models, and aggregate local changes 
(see Table~\ref{table_variance_reduction} for detail). 

On the server side, we consider three adaptive optimizers, Adam, Adagrad, and Yogi, for updating the global model. 
We develop 9 versions of ComFed corresponding to 9 different combinations between the above client-variance reduction mechanisms and adaptive server optimizers. 
Table~\ref{table_intensive} shows the breakdown of FL algorithms, including 7 state-of-the-art algorithms and 9 ComFed variations. It demonstrates which technique to apply for the client and server sides.

\begin{table}
\addtolength{\tabcolsep}{-3pt} 
\begin{tabular}{|l|l|l|l|}

\hline
opt$\_$c	&	Local objective	&	Local upate	&	Aggregation	\\
\hline
Sgd	    &	local loss	&	Sgd	&	averaging	\\
Prox	&	regularized local loss 	&	Sgd	&	averaging	\\
Scaf	&	local loss 	&	corrected Sgd	&	averaging	\\
Nova	&	local loss 	&	Sgd	&	normalized averaging	\\

\hline
\end{tabular}
\caption{Client-variance reduction mechanisms\label{table_variance_reduction}}
\end{table}

\begin{table}
\addtolength{\tabcolsep}{-3pt} 
\begin{tabular}{|ll|l|l|l|l|}

\hline
& & \multicolumn{4}{c|}{Server optimizers}  \\
\cline{3-6}
&		    &	Sgd	        &	Adam	    &	Adagrad	    &	Yogi	\\

\hline
\multirow{5}{*}{\shortstack[c]{Client\\ variance \\reduction}}&	Sgd     &	\it{FedAvg}	    &	\it{FedAdam}	    &	\it{FedAdagrad}	&	\it{FedYogi}	\\

&	Prox    &	\it{FedProx}	    &	ProxAdam	&	ProxAdagrad	&	ProxYogi	\\

&	Scaf	&	\it{Scaffold}	    &	ScafAdam	&	ScafAdagrad	&	ScafYogi	\\

&	Nova	&	\it{FedNova}	    &	NovaAdam	&	NovaAdagrad	&	NovaYogi	\\

	                        
\hline
\end{tabular}
\caption{Decomposition of FL algorithms\label{table_intensive}. State-of-the-art methods are in italic.}
\end{table}

\begin{algorithm*}
\caption{Framework for the baselines and ComFed algorithms\label{alg:our}}

\KwIn{$N, C, T, E, \eta, \eta_g$, client-variance reduction mechanism $opt_c$, server optimizer $opt_s$}
\KwOut {global model $w$}
\BlankLine

\For{round $t = 1, \ldots, T$}{
    Broadcast $w$ (and $c$ if existing) \label{alg:our:broadcast} to a subset $S$ of clients: $|S|=C*N$\\
    
    \For{client $i$ in $S$ \textbf{\textup{in parallel}}}{
    $w_{i} \leftarrow w$,  (and $c_{i} \leftarrow c$ if opt$_c$=scaf)  \label{alg:our:init}\\
    \For{each local epoch and each batch $b_j$ of $D_i$ \label{alg:our:loop_batch}}{
    
        compute mini-batch gradient $g_i$ \\ 
        
    \pushline\dosemic\nonl \textbf{\textit{Local update step}} \\
    \textbf{if} opt$_c$=sgd/nova: \ \ $w_i \leftarrow w_i -\eta g_i $  \label{alg:our:lupdate_Sgd}\\
    \textbf{if} opt$_c$=scaf: \ \ \ \ \ \ \ \ $w_i \leftarrow w_i -\eta (g_i + c - c_i ) \label{alg:our:lupdate_scaf}$      \\
    \textbf{if} opt$_c$=prox: \ \ \ \ \ \ \ $w_i \leftarrow w_i -\eta (g_i + \mu(w_i - w)) $    \\ \label{alg:our:lupdate_Sgd_prox}
    
    \textbf{if} opt$_c$=nova: update $a_i$ \\
     
    }

    \textbf{if} opt$_{c}$=scaf: $c_i^{old} \leftarrow c_i$  ; $c_i \leftarrow \nabla f_i(w,D_i)$ or $c_i \leftarrow c_i - c + \frac{1}{K\eta} (w - w_i)$ \label{alg:our:updatec}\\

    \textbf{if} opt$_c$=sgd/prox: \ \ \ Send back $\Delta_i \leftarrow w_i - w$ to the server         \label{alg:our:send_back1}      \\
    \textbf{if} opt$_c$=nova:  \ \  \ \ \  \ \ \ Send back $\Delta_i \leftarrow w_i - w$ and $a_i$ to the server \label{alg:our:send_back2} \\         
    \textbf{if} opt$_c$=scaf: \ \ \  \ \  \ \  \ \  Send back $(\Delta_i, \Delta c_i) \leftarrow (w_i - w, c_i - c_i^{old})$ to the server \label{alg:our:send_back3} \\
    
    }
    
    \pushline\dosemic\nonl \textbf{\textit{Server aggregation step}} \\
    \textbf{if} opt$_c$=sgd/scaf/prox: \ \ \ $\Delta \leftarrow \sum_{i \in S}{\Delta_i}$ \label{alg:our:aggregation1}\\
    \textbf{if} opt$_c$=nova: \ \ \ \ \ \ \ \ \ \ \ \ \ $\Delta \leftarrow \sum_{i \in S}{p'_i} \lVert a_i \rVert_1 \sum_{i \in S}{p'_i}\frac{\Delta_i}{\lVert a_i \rVert_1}$   \label{alg:our:aggregation2}\\
    \textbf{if} opt$_c$=scaf: \ \ \ \ \ \ \ \ \ \ \ \ \ \ $\Delta_c \leftarrow \sum_{i \in S} p_i' \Delta c_i$ \ \ \ \ \ \ \ \ \ \ \ \ \ \ \ \ \ \  and $c \leftarrow c + \Delta c_i$
     \label{alg:our:aggregation3}\\
    
    \pushline\dosemic\nonl \textbf{\textit{Server update step}} \\

    \textbf{if} opt$_s$=sgd: \ \ \ \ \ \ \ \ \ \ \ $w \leftarrow w + \Delta$ \label{alg:our:gupdate} \\
    \If{ opt$_s$=adam/adagrad/yogi}{
        update momentums $m,v$ according to Equation (1) \\
        update server model $w = \beta_1 w + \eta_g \frac{m}{\sqrt{v} + \varepsilon}$
    } 
    
}
\end{algorithm*}

Algorithm~\ref{alg:our} provides a pseudo--code covering 9 versions of ComFed and 7 baseline methods.
It should be noted that changing the local objective function by a proximal term $\frac{\mu}{2} \lVert {w - w_i} \rVert^2$ (as done in FedProx) will lead to a change of $-\eta \mu \lVert {w - w_i} \rVert$ in the local update. 
So, in the case of Prox, we replace the regularization in the local loss function with an equivalent correction term in the local update (line~\ref{alg:our:lupdate_Sgd_prox}) to present the algorithm implementation systematically. Please refer to~\cite{FedNova} for the exact formula to update $a_i$ when using Sgd, Prox, or Scaf.

\section{Experiments}

\textbf{Non-IID dataset}.
We conduct experiments on the CIFAR-10 image classification challenge that has been widely used as a standard benchmark to evaluate the performance of FL algorithms in prior research. For the Non-IID data setting, we apply Dirichlet distribution $Dir(\alpha)$ to allocate data to clients, where $\alpha$ is a concentration parameter that controls the degree of heterogeneity. The smaller $\alpha$ is, the more skewed the data quantity and label distribution is. Similar to~\cite{FedNova,Adaptive,NIID_silos}, we select $\alpha=0.1$ because the generated data distribution is sufficiently heterogeneous and challenging to evaluate the efficiency of the algorithms on Non-IID data. In this distribution, most of the data samples allocated for each party come from 2 classes.  

\textbf{Model and implementation}. We use the same Cnn model as in~\cite{FedAvg}. 
We develop the algorithms with Flower~\cite{Flower}, an FL framework that provides open-source implementations of several server-side adaptive optimizers, including Adam, Adagrad, and Yogi. 
There are two versions for Scaf corresponding to two approaches for updating the local control variates in Scaffold (see Subsection \ref{subsec_Scaffold}). We use the first version in our experiments since it is more stable than the second one. 

\textbf{Hyper-parameters}.
FL has two main settings: cross-silo (FL between large organizations) and cross-device  (FL between edge devices). 
In cross-silo, the client number is small, around 2-100, and most clients participate in every round. In cross-device, the client number is large, up to a scale of millions, and only a tiny fraction of clients participate in each training round. Primary focus on the more challenging cross-device FL, our experiment uses N = 100
clients (that each client owns a relatively small amount of local data from tens images) with a small client sampling ratio per round C = 0.1 as in the cross-device setting.
Other parameters: training rounds $T=2000$ , local epochs $E=1$, batch size $B=32$, learning rates $\eta=0.01, \eta_g = 0.005$, 
client Sgd with momentum = 0.9, 
and weight decay = 0.0001, 
 server momentum parameters (as in~\cite{Adaptive}) $\beta_1=0.9, \beta_2=0.99$, and Prox's regularization parameter $\mu=0.005$.

\begin{table*}
\addtolength{\tabcolsep}{-3.5pt} 
\begin{center}
\begin{tabular}{|l|l|l|l|l|l|l|l|l|l|l|l|l|l|l|l|l|l|l|l|l|}
\hline
 \diagbox{Algorithm}{Round}	&	100	&	200	&	300	&	400	&	500	&	600	&	700	&	800	&	900	&	1000	&	1100	&	1200	&	1300	&	1400	&	1500	&	1600	&	1700	&	1800	&	1900	&	2000	\\
\hline																																									
\hline

SgdSgd\ \ \ \ \ \ \ \ \ \ \ \ (FedAvg)	&	33.1	&	43.8	&	49.3	&	55.6	&	59.1	&	61.6	&	64.9	&	67.2	&	67.6	&	70.2	&	70.7	&	72.1	&	72.4	&	73.1	&	73.4	&	73.8	&	74.5	&	74.9	&	75.2	&	75.4	\\
SgdAdam\ \ \ \ \ \ \ (FedAdam)	&	35.2	&	43.8	&	46.5	&	49.3	&	51.8	&	53.1	&	54.1	&	55.1	&	56.4	&	57.2	&	58	&	58.6	&	59.2	&	59.9	&	60.2	&	60.7	&	61.2	&	61.7	&	61.7	&	62.5	\\
SgdAdagrad (FedAdagrad)	&	41.6	&	46.8	&	50	&	52.9	&	54.4	&	56.2	&	57.2	&	58.2	&	59.6	&	60.1	&	60.8	&	61.6	&	61.8	&	62.5	&	62.7	&	63.3	&	63.9	&	64.1	&	64.6	&	65.1	\\
SgdYogi\ \ \ \ \ \ \ \ \ \ (FedYogi)	&	43.9	&	52.2	&	56.9	&	61.6	&	63.9	&	66.6	&	68.3	&	69.6	&	70.7	&	71.1	&	\textbf{72.4}	&	72.9	&	73	&	73.4	&	73.6	&	73.8	&	74.2	&	74.3	&	74.4	&	74.7	\\
\hline																																									
NovaSgd \ \ \ \ \ \ \ \ (FedNova)	&	24.5	&	31.2	&	38.2	&	43.3	&	47.2	&	50.7	&	52.9	&	54.3	&	57.2	&	59.9	&	62.5	&	63	&	65.5	&	67.1	&	67.9	&	69.6	&	70.2	&	71.2	&	72.4	&	72.4	\\
NovaAdam	&	24.6	&	35.1	&	40.3	&	44.7	&	48.9	&	52.6	&	57.3	&	57.9	&	60.4	&	62.6	&	64.8	&	65.8	&	68.4	&	68.9	&	70.6	&	70.8	&	71.7	&	72.1	&	73.1	&	73.5	\\
NovaAdagrad	&	26.1	&	32.6	&	38.7	&	41.9	&	47.2	&	49.9	&	51.4	&	55.1	&	57	&	58.7	&	60.6	&	62.7	&	65.7	&	66.8	&	67.7	&	68.7	&	70.5	&	71.5	&	72.1	&	72.4	\\
NovaYogi	&	24.7	&	29.7	&	34.6	&	39.1	&	42.6	&	44.1	&	47.2	&	49	&	52.3	&	55.9	&	58.4	&	59.8	&	62.3	&	63.9	&	65.6	&	66.4	&	68.2	&	68.9	&	70.1	&	70.7	\\
\hline																																									
ProxSgd  \ \ \ \ \ \ \ \ \ (FedProx)	&	34.8	&	41.1	&	48.2	&	53.8	&	57	&	61.9	&	64.3	&	66	&	67.3	&	68.8	&	70.2	&	71.7	&	72.3	&	72.6	&	72.8	&	73.5	&	73.8	&	74.2	&	74.8	&	75.1	\\
ProxAdam	&	37.1	&	43.4	&	47.2	&	49	&	52.2	&	54	&	56.1	&	57.1	&	58.7	&	60.4	&	61.2	&	62	&	62.8	&	63.3	&	64.2	&	64.2	&	65.1	&	65.7	&	66.4	&	66.6	\\
ProxAdagrad	&	40.9	&	46.3	&	49.2	&	52	&	54.4	&	55.6	&	57.1	&	58.4	&	59.3	&	60	&	60.9	&	61.5	&	62	&	62.6	&	63	&	63.6	&	63.9	&	64.4	&	65	&	65.2	\\
ProxYogi	&	\textbf{44.0}	&	\textbf{54.0}	&	\textbf{59.3}	&	\textbf{62.3}	&	\textbf{64.5}	&	\textbf{67.6}	&	\textbf{68.8}	&	\textbf{70.4}	&	\textbf{71.6}	&	\textbf{72.3}	&	\textbf{72.4}	&	\textbf{73.3}	&	\textbf{74.0}	&	\textbf{74.1}	&	\textbf{74.5}	&	\textbf{74.5}	&	\textbf{74.9}	&	\textbf{75.1}	&	\textbf{75.2}	&	\textbf{75.9}	\\
																																									
\hline																																									
ScafSgd  \ \ \ \ \ \ \ \ \ (Scaffold) &	31.1	&	39.2	&	46	&	48.9	&	54.8	&	57	&	59.6	&	61.4	&	64.2	&	65.6	&	66.7	&	67.9	&	68.8	&	69.3	&	70.2	&	70.4	&	71	&	71.6	&	71.9	&	72.1	\\	
ScafAdam	&	35.6	&	42.5	&	44.6	&	48.2	&	51	&	52.8	&	55.4	&	57.2	&	59.1	&	60.1	&	61.2	&	61.8	&	63.1	&	63.3	&	65.3	&	66.2	&	66.2	&	66.7	&	67.4	&	67.7	\\
ScafAdagrad	&	41	&	46.9	&	49.9	&	52.4	&	54.6	&	55.9	&	57.2	&	58	&	58.4	&	59.6	&	60.2	&	60.6	&	60.9	&	61.5	&	62	&	62.3	&	62.8	&	63.3	&	63.5	&	64.2	\\
ScafYogi	&	43.2	&	51.3	&	58	&	60.6	&	63.5	&	65.9	&	67.1	&	68.2	&	69.2	&	70.4	&	71.1	&	71.5	&	72.1	&	72.7	&	72.7	&	73.3	&	73.7	&	73.7	&	74.1	&	74.3	\\
																																									
\hline																																									
\end{tabular}																																							
\caption{The test accuracy ($\%$) on Non-IID CIFAR-10 with 100 clients. The best results are in bold.\label{Cifar10_table}}
\end{center}
\end{table*}

\textbf{Result comparison.}
For each experiment, we run three trials and report the mean accuracy of the best model found after 100, 200, \ldots, and 2000 training rounds, respectively. 
We use Figure~\ref{Cifar10_figure} to visualize the result and  Table~\ref{Cifar10_table} to provide the numbers in detail. 
We see that ProxYogi always achieves the best accuracy. It primarily yields significant improvements over its baselines FedProx and FedAvg, and slight improvements over FedYogi.
At 300 rounds, ProxYogi outperforms FedProx by $11.05\%$, FedYogi by $2.37\%$, and FedAvg by $9.96\%$.
At 1000 rounds, the improvements are respectively $3.51\%, 1.19\%$, and $2.07\%$. Besides,  ScafYogi mitigates the degradation caused by Scaf to produce a reasonable accuracy, sometimes even better than its baselines. For example, at 300 training rounds, ScafYogi achieves $12\%, 1.1\%$, and $8.7\%$ accuracy higher than Scaffold, FedYogi, and FedAvg, respectively (0.580 vs. 0.460, 0.569, and 0.493). 

On the contrary, using Scaf or Nova slows down the convergence. Indeed, Scaf is not compatible with a cross-device setting where the client sampling ratio $C < 1$ because clients cannot maintain local states $c_i$ across rounds.
 Using Adam or Adagrad also decreases accuracy compared to using the vanilla server optimizer Sgd. This degradation could be mitigated when combining Adam with Prox or Scaf. 
 
 In summary, except for ProxYogi and ScafYogi, other ComFed versions $(opt_c$, $opt_s)$ between a client-variance reduction mechanism $opt_c$ and a server-side adaptive optimizer $opt_s$ are not efficient for cross-device FL as we expected. They do not give better accuracy than using $opt_c$ or $opt_s$ alone, or they do not outperform FedAvg.

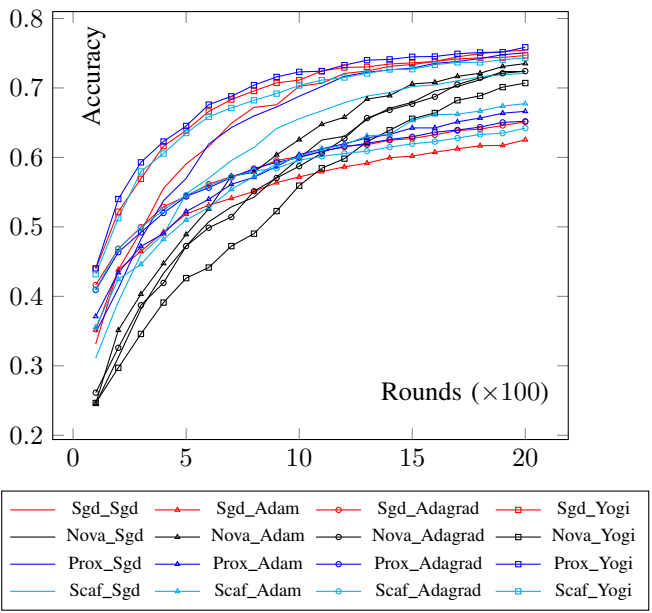
\begin{figure}
\centering
\captionsetup{justification=centering}
\begin{tikzpicture}
  \begin{axis}[ 
        legend columns=4, 
        legend style={font=\scriptsize, at={(-0.1,-0.4)},anchor=south west, column sep=2pt},
        ytick = {0.1, 0.2, 0.3, 0.4, 0.5, 0.6, 0.7, 0.8, 0.9, 1}, 
        ylabel={Accuracy},
        xlabel={Rounds ($\times100$)},
        x label style={at={(axis description cs:0.8,0.25)},anchor=north},
        y label style={at={(axis description cs:0.3,.85)},anchor=south},
      ]
      
    \addplot [mark size=1pt, red] table [x expr = \lineno, y = sgdavg] {Cifar10.dat};    \addlegendentry{Sgd$\_$Sgd}
    \addplot [mark size=1pt, red, mark=triangle] table [x expr = \lineno, y = sgdadam] {Cifar10.dat};    \addlegendentry{Sgd\_Adam}
    \addplot [mark size=1pt, red, mark=o] table [x expr = \lineno, y = sgdadagrad] {Cifar10.dat};    \addlegendentry{Sgd\_Adagrad}
    \addplot [mark size=1pt, red, mark=square] table [x expr = \lineno, y = sgdyogi] {Cifar10.dat};    \addlegendentry{Sgd\_Yogi}    
    
    \addplot [mark size=1pt,  black] table [x expr = \lineno, y = novaavg] {Cifar10.dat};    \addlegendentry{Nova\_Sgd}
    \addplot [mark size=1pt,  black, mark=triangle] table [x expr = \lineno, y = novaadam] {Cifar10.dat};    \addlegendentry{Nova\_Adam}
    \addplot [mark size=1pt,  black, mark=o] table [x expr = \lineno, y = novaadagrad] {Cifar10.dat};    \addlegendentry{Nova\_Adagrad}
    \addplot [mark size=1pt,  black, mark=square] table [x expr = \lineno, y = novayogi] {Cifar10.dat};    \addlegendentry{Nova\_Yogi}   
    
    \addplot [mark size=1pt, blue] table [x expr = \lineno, y = proxavg] {Cifar10.dat};    \addlegendentry{Prox\_Sgd}
    \addplot [mark size=1pt, blue, mark=triangle] table [x expr = \lineno, y = proxadam] {Cifar10.dat};    \addlegendentry{Prox\_Adam}
    \addplot [mark size=1pt, blue, mark=o] table [x expr = \lineno, y = proxadagrad] {Cifar10.dat};    \addlegendentry{Prox\_Adagrad}
    \addplot [mark size=1pt, blue, mark=square] table [x expr = \lineno, y = proxyogi] {Cifar10.dat};    \addlegendentry{Prox\_Yogi}     
    
    \addplot [mark size=1pt, cyan] table [x expr = \lineno, y = scafavg] {Cifar10.dat};    \addlegendentry{Scaf\_Sgd}
    \addplot [mark size=1pt, cyan, mark=triangle] table [x expr = \lineno, y = scafadam] {Cifar10.dat};    \addlegendentry{Scaf\_Adam}
    \addplot [mark size=1pt, cyan, mark=o] table [x expr = \lineno, y = scafadagrad] {Cifar10.dat};    \addlegendentry{Scaf\_Adagrad}
    \addplot [mark size=1pt, cyan, mark=square] table [x expr = \lineno, y = scafyogi] {Cifar10.dat};    \addlegendentry{Scaf\_Yogi}

  \end{axis}
\end{tikzpicture}
\caption{The test accuracy on Non-IID CIFAR-10 with 100 clients \label{Cifar10_figure}}
\end{figure}

\section{Discussion}
Among the baselines, only FedYogi outperforms FedAvg. All other baselines, FedAdam, FedAdagrad, FedProx, FedNova, and Scaffold, yield lower accuracy on the CIFAR-10 challenge with a Non-IID distribution Dirichlet(0.1), unlike~\cite{Adaptive},~\cite{FedProx}, and ~\cite{Scaffold}. This fact is in line with the conclusion in the survey only on client-variance techniques~\cite{NIID_silos} with cross-silo FL that none of the existing FL algorithms outperform others in all cases. The state-of-the-art algorithms outperform FedAvg only in several setting cases. Thus, designing robust FL algorithms for Non-IID data challenges is still a potential research topic.

When both client-variance reduction and adaptive server update techniques speed-up FedAvg, combining them will improve further, like in the case of ProxYogi, where Yogi significantly outperforms FedAvg, and Prox achieves a comparable accuracy with FedAvg. 
On the contrary, when one of the two techniques slows down FedAvg, combing them could reduce the degradation but could not outperform FedAvg. 
The degradation caused by one of the two techniques may be why most of our proposed algorithms are not as efficient as expected. 
In summary, for a given setting consisting of hyper-parameters and data distribution, our proposed combinations make sense when each technique individually beats FedAvg. ProxYogi is the best version of ComFed that outperform all the baselines. 

According to FedAdam, FedAdagrad, FedYogi, and  our ComFed, server momentums are no use to the local updates. This one-way interaction may limit the efficiency of these  algorithms. ~\cite{Mime,FedAdc} show that integrating server momentums into local updates significantly improves the performance of Federated learning in Non-IID data settings because clients can mimic the update of the centralized algorithm. We can apply this idea to improve ComFed as follows:
\begin{itemize}
    \item Local update: $w_i = upd_c(\eta, g_i, mo$)
    \item Server model update: $w = upd_s(\eta_g, \Delta, mo)$
\end{itemize}

where $mo$ denotes server momentums. On the one hand, the new method uses server momentums for local updates instead of just global updates. On the other hand, it determines local updates based on both server momentums and the used client-variance reduction.

\section{Conclusion}
This paper proposed ComFed, a method to address the Non-IID data challenge. It leverages two state-of-the-art techniques dedicated to Non-IID data: client-variance reduction and adaptive global update. The experimental results show that ComFed could improve in some cases for cross-device Federated Learning, i.e., the naive combination of the two techniques could bring advantages compared to using one strategy alone. The relevant case for applying our approach is when both client-variance reduction and adaptive global update techniques, when used alone, outperform FedAvg. 
Integrating server momentums into local updates to mimic the behavior of centralized learning algorithms is one of our future research directions to improve ComFed.

\bibliographystyle{abbrv}
\bibliography{fl}

\begin{thebibliography}{10}

\bibitem{Flower}
D.~J. Beutel, T.~Topal, A.~Mathur, X.~Qiu, T.~Parcollet, and N.~D. Lane.
\newblock Flower: {A} friendly federated learning research framework.
\newblock {\em CoRR}, abs/2007.14390, 2020.

\bibitem{FedDC}
L.~Gao, H.~Fu, L.~Li, Y.~Chen, M.~X. He, and C.-Z. Xu.
\newblock Feddc: Federated learning with non-iid data via local drift
  decoupling and correction.
\newblock In {\em {IEEE} Conference on Computer Vision and Pattern Recognition,
  {CVPR} 2022, New Orleans, Louisiana, June 19-24, 2022}. Computer Vision
  Foundation / {IEEE}, 2022.

\bibitem{FedAvgM}
T.~H. Hsu, H.~Qi, and M.~Brown.
\newblock Measuring the effects of non-identical data distribution for
  federated visual classification.
\newblock {\em CoRR}, abs/1909.06335, 2019.

\bibitem{Mime}
S.~P. Karimireddy, M.~Jaggi, S.~Kale, M.~Mohri, S.~J. Reddi, S.~U. Stich, and
  A.~T. Suresh.
\newblock Mime: Mimicking centralized stochastic algorithms in federated
  learning.
\newblock {\em CoRR}, abs/2008.03606, 2020.

\bibitem{Scaffold}
S.~P. Karimireddy, S.~Kale, M.~Mohri, S.~J. Reddi, S.~U. Stich, and A.~T.
  Suresh.
\newblock {SCAFFOLD:} stochastic controlled averaging for federated learning.
\newblock In {\em Proceedings of the 37th International Conference on Machine
  Learning, {ICML} 2020, 13-18 July 2020, Virtual Event}, volume 119 of {\em
  Proceedings of Machine Learning Research}, pages 5132--5143. {PMLR}, 2020.

\bibitem{NIID_silos}
Q.~Li, Y.~Diao, Q.~Chen, and B.~He.
\newblock Federated learning on non-iid data silos: An experimental study.
\newblock In {\em IEEE International Conference on Data Engineering, {ICDE
  2022}}, 2022.

\bibitem{Moon}
Q.~Li, B.~He, and D.~Song.
\newblock Model-contrastive federated learning.
\newblock In {\em {IEEE} Conference on Computer Vision and Pattern Recognition,
  {CVPR} 2021, virtual, June 19-25, 2021}, pages 10713--10722. Computer Vision
  Foundation / {IEEE}, 2021.

\bibitem{FedProx}
T.~Li, A.~K. Sahu, M.~Zaheer, M.~Sanjabi, A.~Talwalkar, and V.~Smith.
\newblock Federated optimization in heterogeneous networks.
\newblock In I.~S. Dhillon, D.~S. Papailiopoulos, and V.~Sze, editors, {\em
  Proceedings of Machine Learning and Systems 2020, MLSys 2020, Austin, TX,
  USA, March 2-4, 2020}. mlsys.org, 2020.

\bibitem{qFFL}
T.~Li, M.~Sanjabi, A.~Beirami, and V.~Smith.
\newblock Fair resource allocation in federated learning.
\newblock In {\em 8th International Conference on Learning Representations,
  {ICLR} 2020, Addis Ababa, Ethiopia, April 26-30, 2020}. OpenReview.net, 2020.

\bibitem{FedAvg_NIID}
X.~Li, K.~Huang, W.~Yang, S.~Wang, and Z.~Zhang.
\newblock On the convergence of fedavg on non-iid data.
\newblock In {\em 8th International Conference on Learning Representations,
  {ICLR} 2020, Addis Ababa, Ethiopia, April 26-30, 2020}. OpenReview.net, 2020.

\bibitem{VLRSGD}
X.~Liang, S.~Shen, J.~Liu, Z.~Pan, E.~Chen, and Y.~Cheng.
\newblock Variance reduced local {SGD} with lower communication complexity.
\newblock {\em CoRR}, abs/1912.12844, 2019.

\bibitem{FedAvg}
B.~McMahan, E.~Moore, D.~Ramage, S.~Hampson, and B.~A. y~Arcas.
\newblock Communication-efficient learning of deep networks from decentralized
  data.
\newblock In A.~Singh and X.~J. Zhu, editors, {\em Proceedings of the 20th
  International Conference on Artificial Intelligence and Statistics, {AISTATS}
  2017, 20-22 April 2017, Fort Lauderdale, FL, {USA}}, volume~54 of {\em
  Proceedings of Machine Learning Research}, pages 1273--1282. {PMLR}, 2017.

\bibitem{FedAdc}
E.~Ozfatura, K.~Ozfatura, and D.~G{\"{u}}nd{\"{u}}z.
\newblock Fedadc: Accelerated federated learning with drift control.
\newblock In {\em {IEEE} International Symposium on Information Theory, {ISIT}
  2021, Melbourne, Australia, July 12-20, 2021}, pages 467--472. {IEEE}, 2021.

\bibitem{Adaptive}
S.~J. Reddi, Z.~Charles, M.~Zaheer, Z.~Garrett, K.~Rush, J.~Kone{\v{c}}n{\'y},
  S.~Kumar, and H.~B. McMahan.
\newblock Adaptive federated optimization.
\newblock In {\em 9th International Conference on Learning Representations,
  {ICLR} 2021, Virtual Event, Austria, May 3-7, 2021}. OpenReview.net, 2021.

\bibitem{FedNova}
J.~Wang, Q.~Liu, H.~Liang, G.~Joshi, and H.~V. Poor.
\newblock Tackling the objective inconsistency problem in heterogeneous
  federated optimization.
\newblock In H.~Larochelle, M.~Ranzato, R.~Hadsell, M.~Balcan, and H.~Lin,
  editors, {\em Advances in Neural Information Processing Systems 33: Annual
  Conference on Neural Information Processing Systems 2020, NeurIPS 2020,
  December 6-12, 2020, virtual}, 2020.

\bibitem{SlowMo}
J.~Wang, V.~Tantia, N.~Ballas, and M.~G. Rabbat.
\newblock Slowmo: Improving communication-efficient distributed {SGD} with slow
  momentum.
\newblock In {\em 8th International Conference on Learning Representations,
  {ICLR} 2020, Addis Ababa, Ethiopia, April 26-30, 2020}. OpenReview.net, 2020.

\bibitem{FedAvg_NIID_shareddata}
Y.~Zhao, M.~Li, L.~Lai, N.~Suda, D.~Civin, and V.~Chandra.
\newblock Federated learning with non-iid data.
\newblock {\em CoRR}, abs/1806.00582, 2018.

\end{thebibliography}
\end{document}